\documentclass[letterpaper,11pt]{article}

\usepackage{header}
\usepackage[margin=1in]{geometry}
\usepackage{textcomp,graphicx,enumerate}
\usepackage{booktabs}
\usepackage[numbers]{natbib}
\usepackage{subcaption}
\usepackage{multirow}
\usepackage{comment}

\title{Drag-guided diffusion models for vehicle image generation}
\author{%
  Nikos Ar\'echiga, Frank Permenter, Chenyang Yuan\footnote{Authors listed alphabetically}\\
  Toyota Research Institute\\
  \texttt{firstname.lastname@tri.global} \\
  \\
  Binyang Song\\
  Massachusetts Institute of Technology\\
  \texttt{binyangs@mit.edu}
}

\begin{document}
\maketitle
\begin{abstract}
  Denoising diffusion models trained at web-scale have revolutionized
  image generation.  The application of these tools to engineering design is an
  intriguing possibility, but is currently limited by their inability to parse
  and enforce concrete engineering constraints. In this paper, we take a step
  towards this goal by proposing physics-based guidance, which enables
  optimization of a performance metric (as predicted by a surrogate model)
  during the generation process.  As a proof-of-concept, we add drag guidance to
  Stable Diffusion, which allows this tool to generate images of novel vehicles
  while simultaneously minimizing their predicted drag coefficients.
\end{abstract}

\section{Introduction}
\begin{figure}
\centering
\begin{subfigure}{0.45\textwidth}
\centering
  \includegraphics[width=0.5\textwidth]{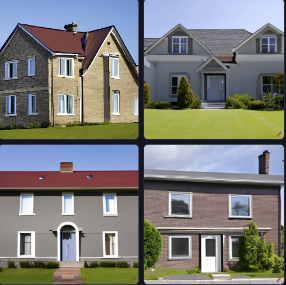}
  \caption{``a house with two windows''}
\label{fig:badHouse}
\end{subfigure}%
\begin{subfigure}{0.45\textwidth}
\centering
\includegraphics[width=0.5\textwidth]{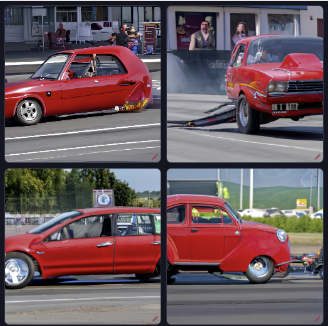}
\caption{``a car with a drag coefficient of 0.5''}
\label{fig:badDrag}
\end{subfigure}
\caption{Illustrates the inability of text-guidance to realize quantitative design specifications.
Images were generated by \texttt{craiyon.com}}
\end{figure}

Diffusion models have produced state of the art results for high-resolution
image generation. One powerful aspect of diffusion models is the incorporation
of \emph{text-based guidance}, which steers the image
generation process towards an output that matches a user-provided description.
Such guidance is  implemented in DALL-E 2 \cite{ramesh2022hierarchical}, Imagen
\cite{saharia2022photorealistic} and Stable Diffusion~\cite{rombach2022high}.
Given their capabilities, it is natural to ask how current image generation
tools can be applied to engineering design, and, in particular, if
text-guidance is sufficient to specify quantitative engineering constraints.
Simple experiments reveal significant challenges.  For example, given a text
prompt of ``a house with \emph{two} windows'', a text-guided diffusion model
produces a house with windows, but ignores the quantitative constraint
on the specified \emph{number} of windows (\Cref{fig:badHouse}).  The prompt ``a car with a drag
coefficient of 0.5'' is also completely misinterpreted (\Cref{fig:badDrag}),
showing an inability of the generation process to even correctly parse the constraint.

To successfully incorporate engineering constraints
into existing image generation tools, it is clear that new guidance approaches
must be developed. In this paper, we implement \emph{physics-based}
guidance that simultaneously estimates and optimizes
the performance of the generated object.
As a proof-of-concept, we implement \emph{drag} optimization of \emph{vehicles},
but our approach can be adapted to any design task
that admits a \emph{surrogate model} of performance
with the following characteristics:
\begin{enumerate}
\item The input is an image, as produced by a diffusion process for image generation.
\item The model is robust to distributional shifts in the input.
\item The model is differentiable, allowing for gradient-based
optimization of the output.
\end{enumerate}
For the task of vehicle drag optimization, we train a surrogate model that meets
these specifications using the dataset~\cite{song2023}, which consists of
two-dimensional (2D) vehicle renderings labeled with drag coefficients
calculated from fluid-dynamics simulations.  Aiming for robustness to
distributional shifts, we build our model using feature-space embeddings from
widely used object recognition networks~\cite{radford2021learning, he2016deep,
  dosovitskiy2020image}.  Finally, our model is implemented using a deep-neural
network and is hence differentiable.

\begin{figure}
  \centering
  \includegraphics[width=0.8\textwidth]{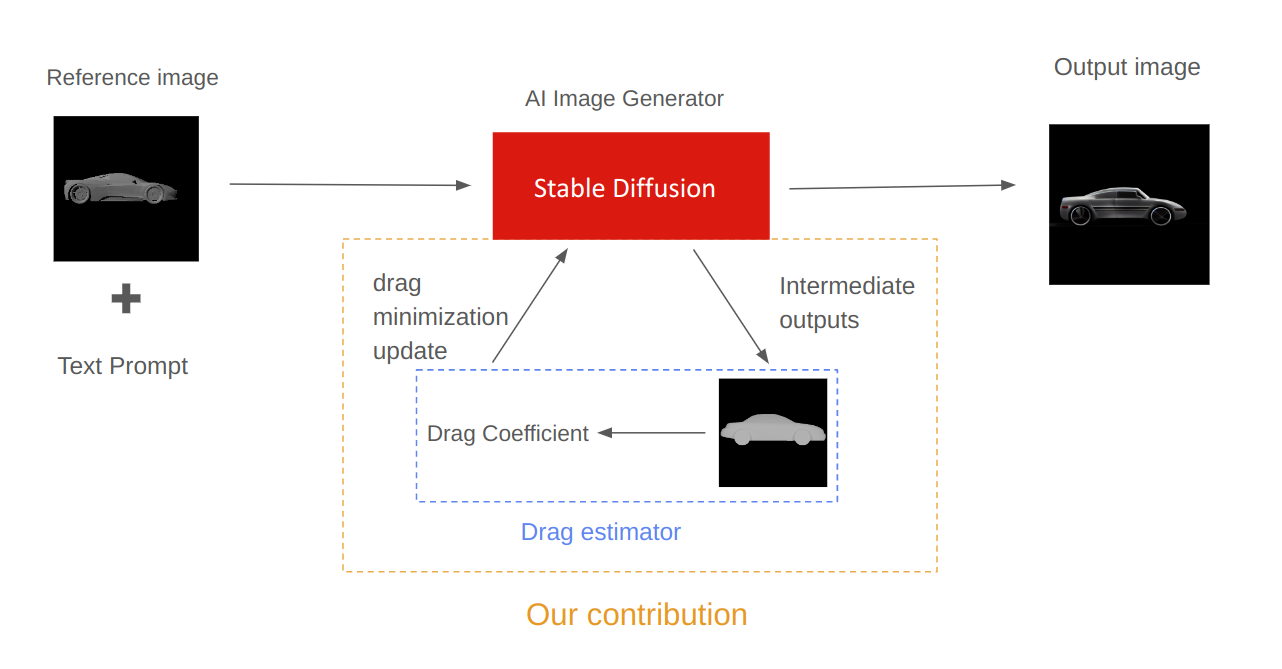}
  \caption{Our pipeline for incorporating drag minimization into diffusion
    models.}~\label{fig:pipeline}
\end{figure}

\begin{figure}[t]
  \centering
  \begin{tabular}{cc}
    \textbf{Baseline} & \textbf{Drag optimized} \\[0.3cm]
    \includegraphics[width=0.2\textwidth]{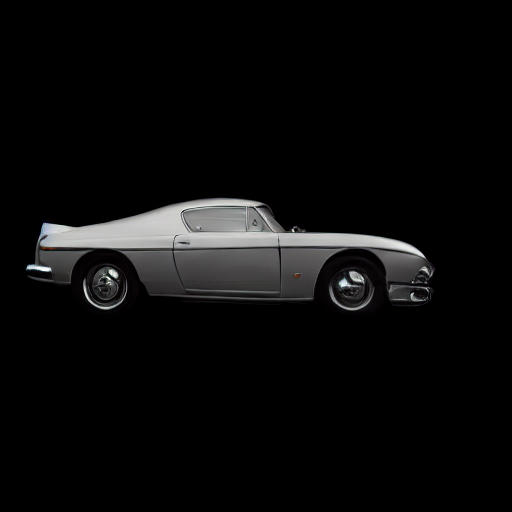} &
    \includegraphics[width=0.2\textwidth]{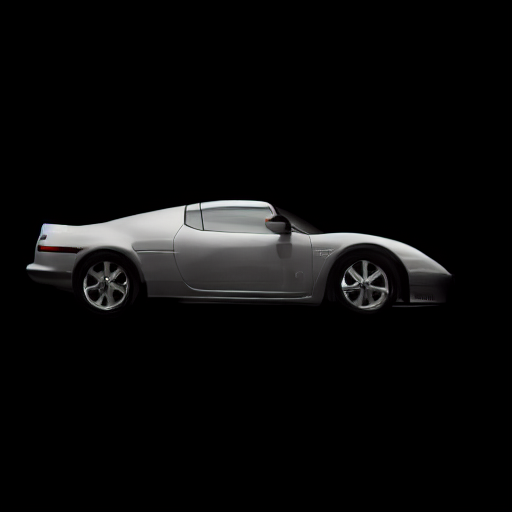} \\
    \includegraphics[width=0.2\textwidth]{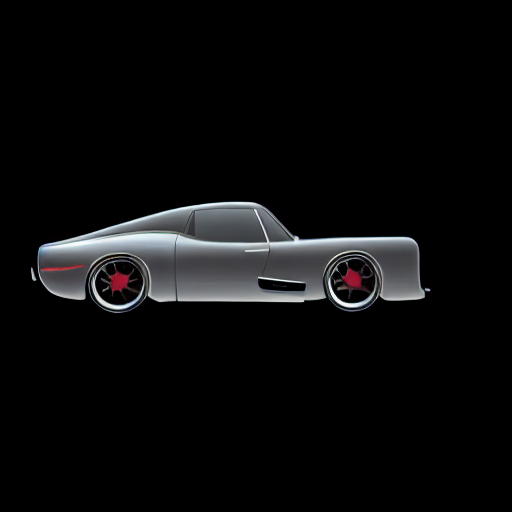} &
    \includegraphics[width=0.2\textwidth]{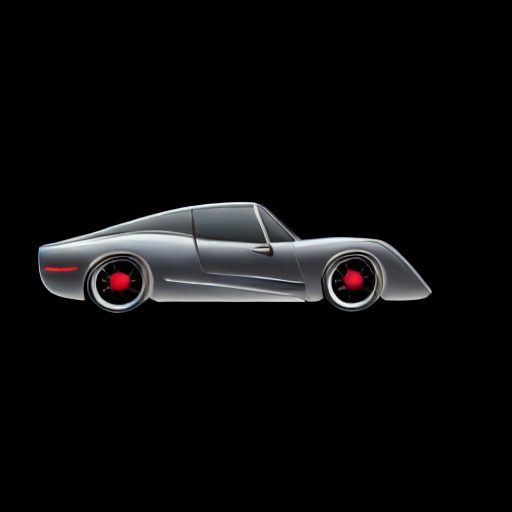}\\
  \end{tabular}
  \begin{tabular}{cc}
    \textbf{Baseline} & \textbf{Drag optimized} \\[0.3cm]
    \includegraphics[width=0.2\textwidth]{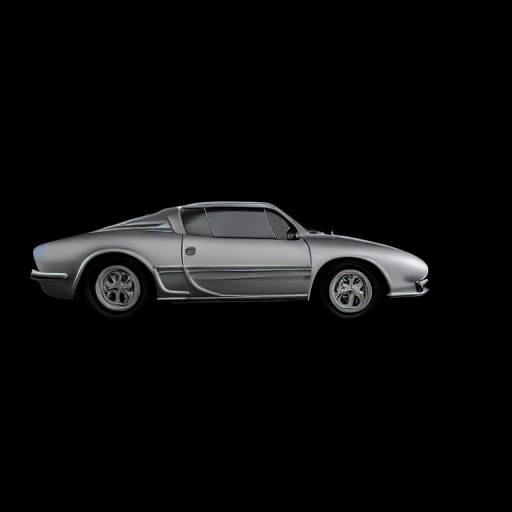} &
    \includegraphics[width=0.2\textwidth]{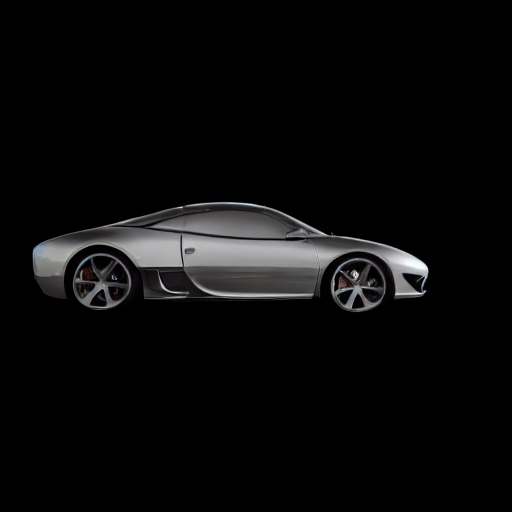} \\
    \includegraphics[width=0.2\textwidth]{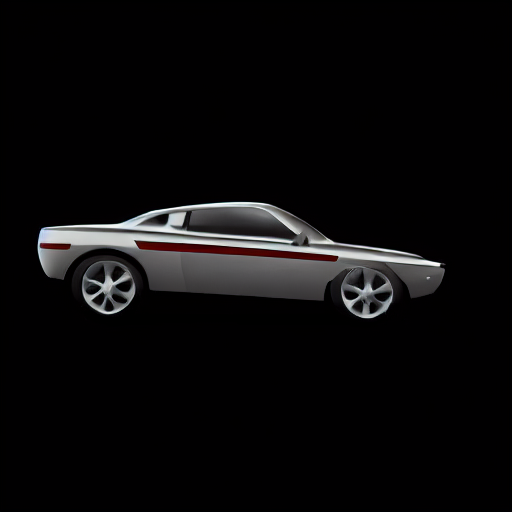} &
    \includegraphics[width=0.2\textwidth]{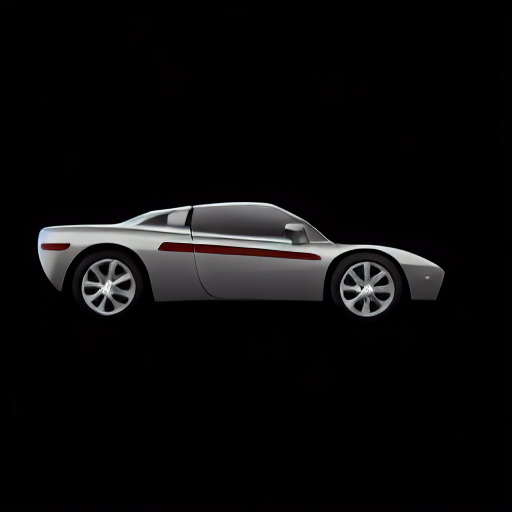}
  \end{tabular}
  \caption{Outputs of Stable Diffusion  with and without our drag-guidance technique.}
  \label{fig:outputs}
\end{figure}

Leveraging our surrogate model, we integrate \emph{drag guidance}\footnote{In
  this paper, we use ``drag'' and ``drag coefficient'' interchangeably.}  into
Stable Diffusion (\Cref{fig:pipeline}),
and show that this integrated system can generate car images that both match a
text prompt and  minimize predicted drag.  We also find that drag guidance can
produce more aesthetically streamlined outputs.
\Cref{fig:outputs} contains illustrative examples of our results
compared to baseline images generated by an unmodified version of Stable Diffusion.

\section{Related Work}
\paragraph{Guided diffusion}
Guidance aims to minimize an auxiliary loss function $\phi(x)$ during the
image generation process~\cite{bansal2023universal}.  This is done by
incorporating the gradient of $\phi(x)$ into each generation step. Guidance
techniques can be differentiated by the type of loss function.  In
\emph{classifier-based} guidance \cite{dhariwal2021diffusion}, the loss function
$\phi(x)$ is the logit probability of a specified class in a pre-trained
classifier.  The aim of this guidance is to steer the output towards this class.
In \emph{text-based} guidance, $\phi(x)$ is a similarity score computed in a
latent space, e.g., the cosine similarity between the CLIP embedding of the
image $x$ and a given text-prompt~\cite{radford2021learning, crowson_2021, Nichol2021GLIDE:Models, Kim2021DiffusionCLIP:Manipulation}.
 Guidance can also be used to solve linear equations $Ax=b$
by setting $\phi(x) = \|b-Ax\|^2$. This setup generalizes image inpainting,
compressed sensing, and super-resolution~\cite{chung2022improving,chung2022diffusion}.  Alternatively,
one can implement \emph{classifier-free} guidance, in which gradients of
 $\phi(x)$ are effectively learned during training. Such approaches have
been used for text guidance~\cite{ho2022classifier},
inpainting~\cite{rombach2022high} and single-view reconstruction
\cite{liu2023zero}.
\paragraph{Diffusion for design}
Prior applications of diffusion to engineering design
include~\cite{giannone2023diffusing, maze2022topodiff}, which train \emph{new}
diffusion models for a specific design task.  In contrast, our work focuses on
adapting an \emph{existing} diffusion model (Stable Diffusion), aiming to
leverage the web-scale dataset used for its training.
\paragraph{Drag optimization and machine learning}

Drag optimization is a classical problem that can be solved using
PDE-optimization techniques.  These approaches require accurate solutions of the
Navier-Stokes equations, which are computationally expensive to obtain and
sensitive to shape representation. Circumventing these difficulties using
ML-techniques is an active area of research.  One thread of research replaces
direct solution of the Navier-Stokes equations with a \emph{surrogate
  model}. These models are trained using computational fluid-dynamics (CFD)
simulations and either predict pressure or velocity
fields~\cite{umetani2018learning, remelli2020meshsdf, allen2022physical,
  durasov2021debosh, baque2018geodesic,abbas2022geometrical}, and/or a
performance measure (e.g., drag coefficient) directly from model
parameters~\cite{song2023, baque2018geodesic}. Different model architectures
have been used, including Gaussian process
regression~\cite{umetani2018learning}, graph neural
networks~\cite{allen2022physical, durasov2021debosh}, geodesic convolutional
neural networks \cite{baque2018geodesic,abbas2022geometrical}, variational
autoencoders (VAEs) \cite{saha2021exploiting}, and
UNets~\cite{thuerey2020deep,jacob2021deep}.  Our work continues this thread and
develops a surrogate model for drag coefficient estimation using \emph{linear
  regression} on features extracted from pretrained neural-networks.  Another
thread concerns novel shape representations that maintain end-to-end
differentiability when cascaded with a surrogate model. Such representations
include Deep Implicit Fields~\cite{remelli2020meshsdf},
PolyCubes~\cite{umetani2018learning}, and 2D
renderings~\cite{rosset2023interactive, thuerey2020deep,jacob2021deep,
  song2023}.  Our work uses 2D renderings proposed by~\cite{song2023}, which
capture the point-wise depth and normal information of three dimensional (3D)
shapes using RGB color channels of pixel data. This image representation allows
us to easily implement a differentiable surrogate-model that can predict drag
coefficients directly from intermediate Stable Diffusion outputs.

\section{Denoising Diffusion Models}
Diffusion models are a class of generative models inspired by the
physical process of diffusion. These models are trained
by adding progressively larger amounts of Gaussian noise to a given data set,
mimicking a diffusion process.  Sampling reverses this process through \emph{denoising}.

\subsection{Training}
Diffusion models estimate
a \emph{noise vector} $\epsilon \in \mathbb{R}^n$
from a given $y \in \mathbb{R}^n$ and \emph{noise level} $\sigma > 0$
such that $y = x + \sigma \epsilon$
approximately holds for some $x$ in the training set.
The learned estimator, denoted
$\epsilon_{\theta} : \mathbb{R}^n \times \mathbb{R}_+ \rightarrow \mathbb{R}^n$,
is called a \emph{denoiser}. The trainable parameters, denoted jointly by  $\theta \in \mathbb{R}^m$,
are found by (approximately) minimizing
\begin{align}\label{eq:trainingLoss}
  L(\theta): = \Ex_{x, \sigma, \epsilon} \norm{\epsilon_\theta(x +  \sigma \epsilon, \sigma) - \epsilon}^2
\end{align}
when $x$ is drawn from the training-set distribution, $\sigma$ is drawn
uniformly from a finite set of positive numbers, and $\epsilon$ is drawn from a Gaussian
distribution $\mathcal{N}(0, I)$.

Current state-of-the-art diffusion models modify this training process in a few
ways. Models supporting classifier-free text guidance add text embeddings (from
CLIP, for example) to the denoiser input~\cite{radford2021learning}.  Models
supporting high-resolution image synthesis implement \emph{latent
  diffusion}~\cite{rombach2022high}, i.e., the variable $x$ is not an element of
pixel space, but, rather, an element of a lower-dimensional latent space defined
by a VAE.

\subsection{Sampling}\label{sec:sampling}

Let $\{\sigma_t\}^T_{t=0}$ denote a set of noise levels $\sigma$ used in training and assume
that $\sigma_t > \sigma_{t-1}$. The sequence $\sigma_t$  is used by \emph{sampling algorithms}
to construct novel images using a trained denoiser $\epsilon_\theta$.
We use the DDIM sampler~\cite{song2020denoising} which, given $x_T \sim \mathcal{N}(0, \sigma_T^2 I)$,
generates an image $x_0$  via the following recursion\footnote{We note~\cite{song2020denoising} presents this recursion in coordinates
$z_t := \sqrt{\alpha_t} x_t$ where $\alpha_t$ satisfies $\sigma^2_t = (1-\alpha_t)/\alpha_t$.}
\begin{align} \label{eq:ddim-update-step}
  x_{t-1} = x_t - (\sigma_t - \sigma_{t-1}) \epsilon_\theta(x_t, t).
\end{align}
Classifier-free text-guidance~\cite{ho2022classifier}
executes this recursion using a modified denoiser
$\epsilon_\theta(x_t, t, y)$ that takes a reference latent embedding $y$ (e.g., from CLIP)
as an optional third input.
Concretely, it constructs $\epsilon_\theta(x_t, t)$ via
\begin{align} \label{eq:cond-guidance}
  \epsilon_\theta(x_t, t) = (1-w) \epsilon_\theta(x_t, t, \varnothing)
  + w (\epsilon_\theta(x_t, t, y))
\end{align}
where $w$ is a weighing factor and $\varnothing$ denotes the empty string.

\section{Method and Implementation}
In this section, we first describe the construction of a surrogate model for
drag coefficient estimation.  We then describe our technique for drag
  guidance, which incorporates gradients of this model into the sampling
process.

\subsection{A surrogate model for drag estimation}\label{sec:drag_estimator}
\begin{figure}
  \centering
  \includegraphics[width=0.8\textwidth]{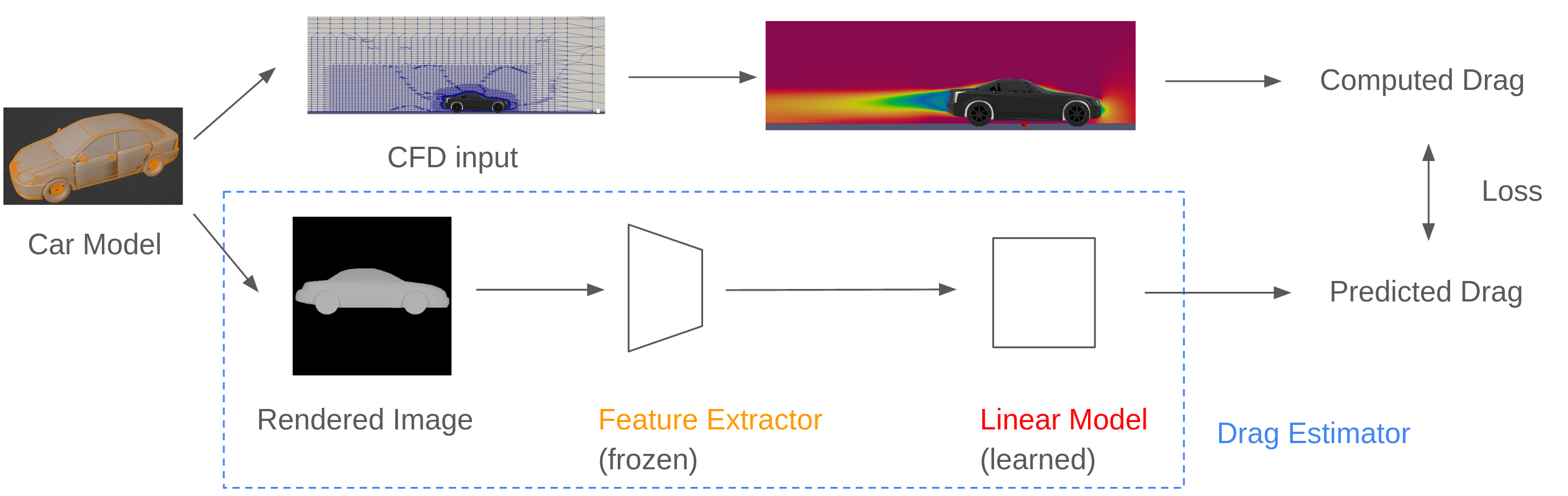}
  \caption{Training the drag estimator}
  \label{fig:drag-estimator}
\end{figure}

To estimate drag, we train a surrogate model that predicts drag coefficients
from side-view vehicle images (\Cref{fig:drag-estimator}).  Training uses the
dataset~\cite{song2023}, which consists of 2D vehicle renderings labeled with
drag coefficients calculated from CFD simulations.  Our model consists of a
trainable linear-layer attached to a frozen feature extractor, a
well-established technique in transfer learning for improving robustness to
out-of-distribution shifts \cite{kumar2022fine}. More details of the training
process can be found in \Cref{sec:details}.  \Cref{table:features} compares
model performance for different feature extractors, specifically, CLIP
\cite{radford2021learning}, ResNet \cite{he2016deep}, Vision Transformers
\cite{dosovitskiy2020image}, and random convolutions \cite{rahimi2007random}.

\begin{figure}
\centering
\begin{subfigure}[t]{0.4\textwidth}
  \centering
  \caption{Naive drag optimization on pixel space}
  \label{fig:drag-optimization-naive}
  \includegraphics[width=0.7\textwidth]{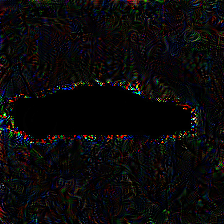}
\end{subfigure}%
\begin{subfigure}[t]{0.6\textwidth}
  \centering
  \def\svgwidth{\textwidth}
  \caption{Drag optimization interleaved with denoising}
  \label{fig:drag-optimization}
  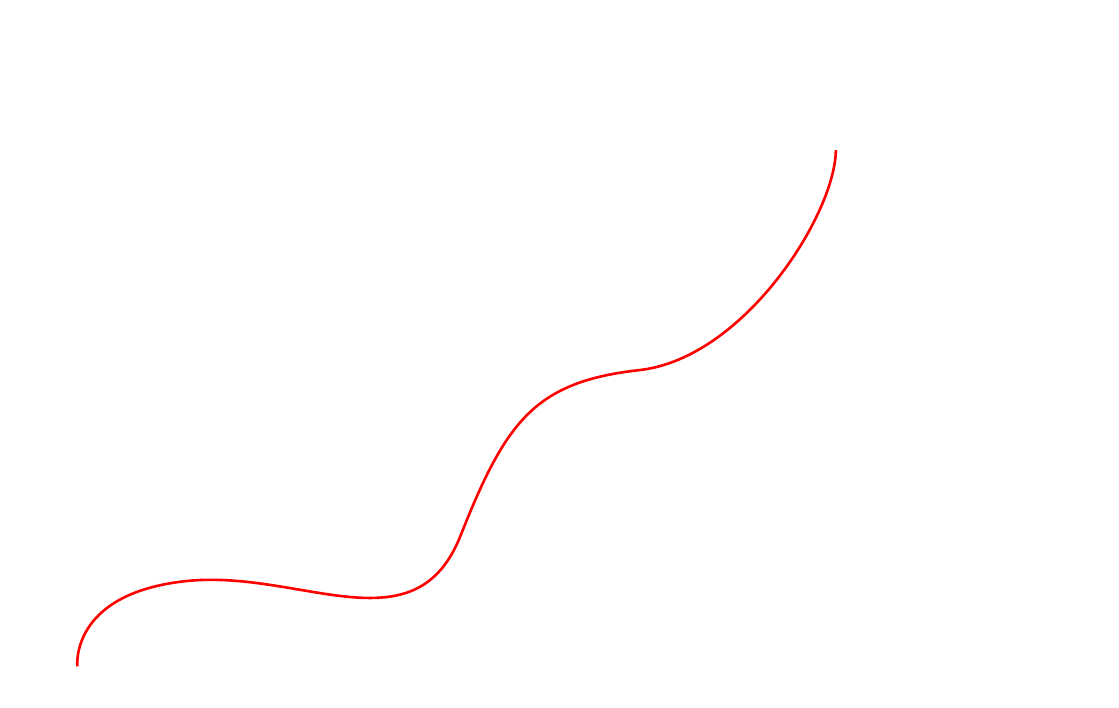%
\end{subfigure}
\caption{An illustration of how drag guidance works.
Noisy images produced by naively minimizing the surrogate model (\Cref{fig:drag-optimization-naive})
are avoided by the inclusion of denoising steps (\Cref{fig:drag-optimization}).}
\end{figure}

Note that the surrogate model alone is not
sufficient to perform drag optimization: starting with an image of a car and
minimizing the predicted drag in pixel space produces results similar to
\Cref{fig:drag-optimization-naive}. We remedy this using guided diffusion,
which will interleave drag minimization steps with denoising steps.

\begin{table}
  \centering
  \begin{tabular}{ccccc}
    \toprule
    Features & Dimension & Train $R^2$ & Test $R^2$ & Test MSE \\
    \midrule
    CLIP & 512 & 0.639 & 0.545 & 0.00251 \\
    ViT & 768 & 0.639 & 0.549 & 0.00249 \\
    ResNet & 2048 & 0.630 & 0.509 & 0.00271 \\
    Random & 2560 & 0.640 & 0.554 & 0.00246 \\
    \bottomrule
  \end{tabular}
  \caption{Performance of drag estimator with different feature extractors.}~\label{table:features}
\end{table}

\subsection{Drag-guided diffusion}
Let $\phi_\text{drag}: \R^{C \times H \times W} \rightarrow \R$
denote our surrogate model.
To implement drag-guided diffusion, we
follow~\cite[Algorithm 2]{dhariwal2021diffusion}
and modify the sampling step~\eqref{eq:ddim-update-step} to
\begin{align}\label{eq:conditioning}
  x_{t-1} &= x_t - (\sigma_t - \sigma_{t-1}) (\epsilon_\theta(x_t, t) + \eta_t \nabla\phi_\text{drag}(\hat{x}_0^t)),
\end{align}
where $\eta_t$ denotes an iteration-dependent weight and $\hat{x}_0^t$ denotes
the denoised output, defined as
$\hat{x}_0^t = x_t - \sigma_t \epsilon_\theta(x_t, t)$.
Per~\cite{permenter2023interpreting}, we can interpret construction of $\hat{x}_0^t$
as approximate projection onto the image manifold. The iterations~\eqref{eq:conditioning}
in turn can be reinterpreted as a damped form of projected gradient descent, an algorithm
that alternates between projection and gradient steps.
Precisely,~\eqref{eq:conditioning} implicitly constructs a shadow iterate $\hat x_\text{drag}^t$ using a projected-gradient
descent iteration, and then dampens the update to $\hat x_\text{drag}^t$ using the
quantity $\alpha_t:= 1-\sigma_{t-1}/\sigma_t$.
Consider the following.
\begin{prop}
The iterations~\eqref{eq:conditioning} are equivalent to
\begin{align}
\hat x^t_0 = x_t - \sigma_t \epsilon_\theta(x_t, t)~\label{eq:denoise} \\
\hat x_\text{drag}^t = \hat x^t_0 - \gamma_t \nabla\phi_\text{drag}(\hat x^t_0) \nonumber\\
x_{t-1} = (1-\alpha_t)  x_t + \alpha_t  \hat x_\text{drag}^t \nonumber
\end{align}
where $\alpha_t = 1 - \frac{\sigma_{t-1}}{\sigma_t}$
and $\gamma_t    = \sigma_t \eta_t$.
\begin{proof}
Expanding and simplifying the last equation gives
\begin{align*}
x_{t-1} &= (1-\alpha_t)  x_t + \alpha_t  [ x_t - \sigma_t \epsilon_\theta(x_t, t) - \gamma_t \nabla\phi_\text{drag}(\hat x^t_0)  ]\\
        &=   x_t + \alpha_t [ \sigma_t \epsilon_\theta(x_t , t) - \gamma_t \nabla\phi_\text{drag}(x^t_0)  ].
\end{align*}
Equating with~\eqref{eq:conditioning} shows that
$\sigma_t - \sigma_{t-1} = \alpha_t  \sigma_t$
and $(\sigma_t - \sigma_{t-1}) \eta_t = \alpha_t \gamma_t$.
Rearranging proves the claim.
\end{proof}
\end{prop}
In total, we can interpret construction of $\hat x^t_0$ as stepping towards
the image manifold, and construction of $\hat x_\text{drag}^t$ as a drag-minimization step.
This is illustrated by \Cref{fig:drag-optimization}.

\subsection{Implementation details} \label{sec:details}
To train our model, we used side-view renderings from the dataset~\cite{song2023}.
For each item in this dataset, we randomly generated 10 new training examples
using horizontal flips, vertical shifts (up to 25 pixels) and
jitter (up to five percent) in brightness, contrast, saturation and hue.
 The renderings were also resized to 224 by 224 pixels to
match the input size of the pretrained feature extractors, which we obtained
from the following Huggingface repositories:
\[
\footnotesize
\texttt{openai/clip-vit-base-patch32}, \;\;
\texttt{google/vit-base-patch16-224-in21k}, \;\; \texttt{microsoft/resnet-50}.
\]
We also compared performance to random convolutional features, which we
constructed using a kernel size of five and weights drawn from a standard normal
distribution. At the output of these convolutions, we applied a bias of +2, a
ReLU nonlinearity, and a pooling operation using a 55 by 55 window.  After
extracting the features, we use ridge regression with regularization parameter
$\lambda=100$ and
$\lambda=10$, for the pretrained and random features respectively, to train the
last linear layer.

Finally, drag guidance was incorporated into a checkpoint of Stable Diffusion
available at \texttt{runwayml/stable-diffusion-v1-5}, with
$\eta_t = \eta_0/\sqrt{1+1/\sigma_t^2}$, and $\eta_0 = 400$. During sampling, we
fix a reference image $x_0$, which is a side-view rendering of a car, and set
$T=80$. Then we let $x_T = x_0 + \sigma_T \epsilon$, where
$\epsilon \sim \mathcal{N}(0, I)$, a commonly used technique in image-to-image generation
\cite{meng2021sdedit}. We also fix a text prompt and use a scale of 7.5 for
classifier-free text guidance.

\section{Experimental Results}
\subsection{Generation of novel designs}
\begin{figure}
  \centering
  \begin{tabular}{ccc}
    \textbf{Baseline} & \textbf{Drag optimized} & \textbf{Estimated Drag} \\[0.3cm]
    \includegraphics[width=0.25\textwidth]{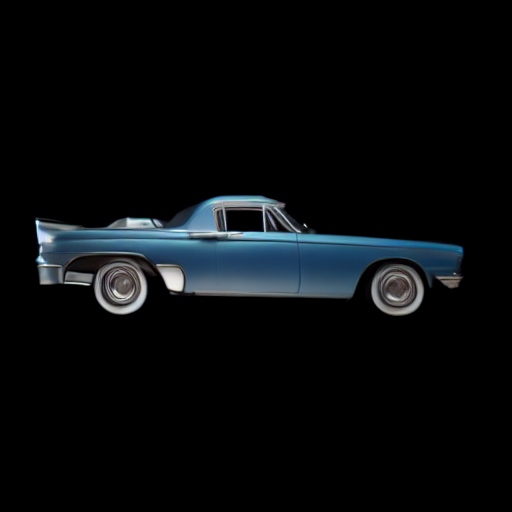}&\includegraphics[width=0.25\textwidth]{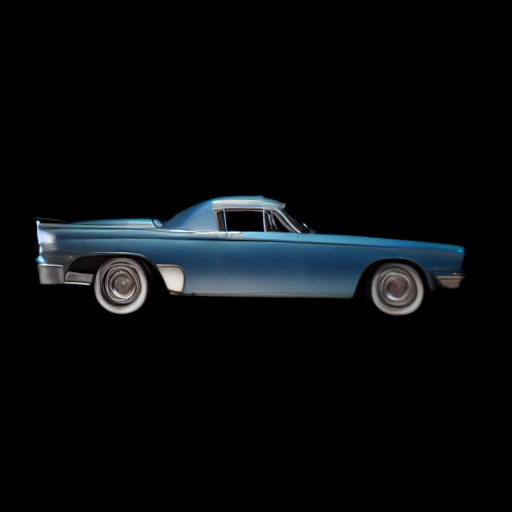}&\includegraphics[width=0.3\textwidth]{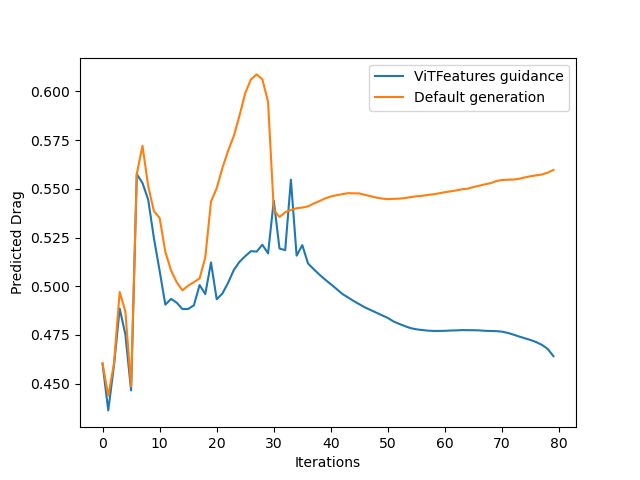}\\ \includegraphics[width=0.25\textwidth]{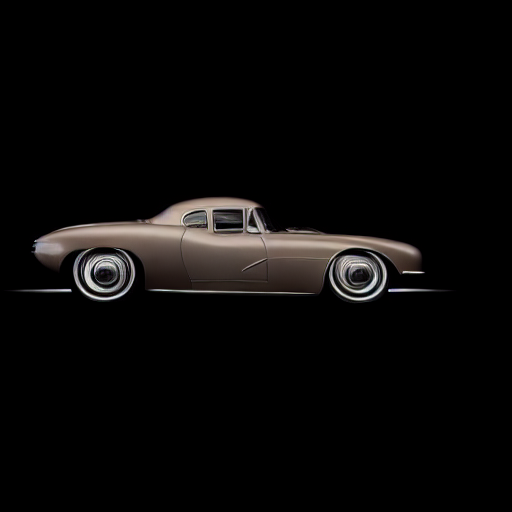}&\includegraphics[width=0.25\textwidth]{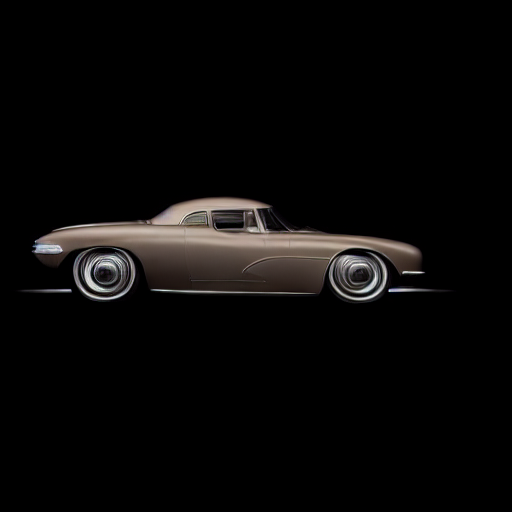}&\includegraphics[width=0.3\textwidth]{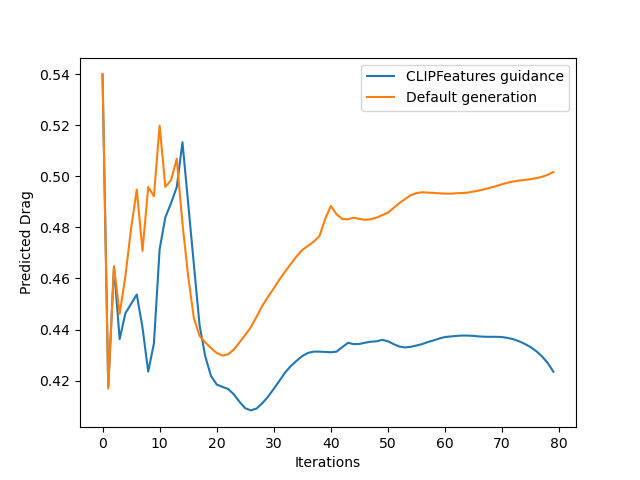}\\
  \end{tabular}
  \caption{Outputs of our image generation pipeline with drag optimization,
    compared to baseline images with the drag optimization turned off. }
  \label{fig:outputs_more}
\end{figure}

We first consider generation of novel vehicle images with drag
guidance. Compared to baseline generation, we find that drag guidance can change
the generated image in subtle ways (e.g. artifact removal or surface smoothing
as in \Cref{fig:outputs_more}) or through significant redesigns (e.g, adding a
roof to a convertible as in \Cref{fig:outputs_hos}). In all cases, however, a
coherent vehicle shape is preserved, as is the artistic style, showing that
text-based guidance can work in concordance with drag optimization. Sample
results are shown in \Cref{fig:outputs} and \Cref{fig:outputs_more}.
\Cref{fig:outputs_more} also confirms that drag guidance indeed decreases
predicted drag.  \Cref{sec:additional_ex} contains more examples.

\subsection{Drag-guided redesign}
\begin{figure}
  \centering
  \begin{tabular}[t]{rcc}
    \textbf{Reference} & \scalebox{-1}[1]{\includegraphics[width=0.35\textwidth]{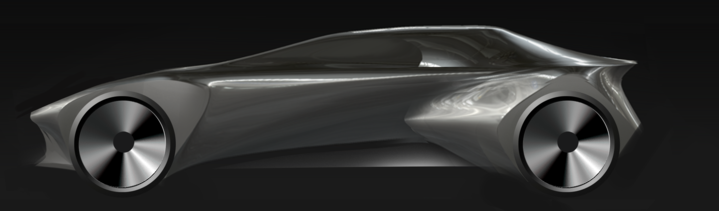}} &   \multirow{3}{*}{\includegraphics[width=0.5\textwidth]{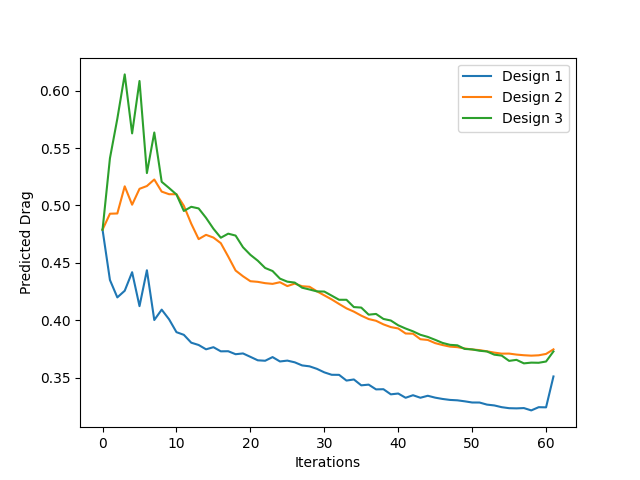}}\\
    \textbf{Design 1} & \includegraphics[width=0.35\textwidth,trim={0 180px 0 180px},clip]{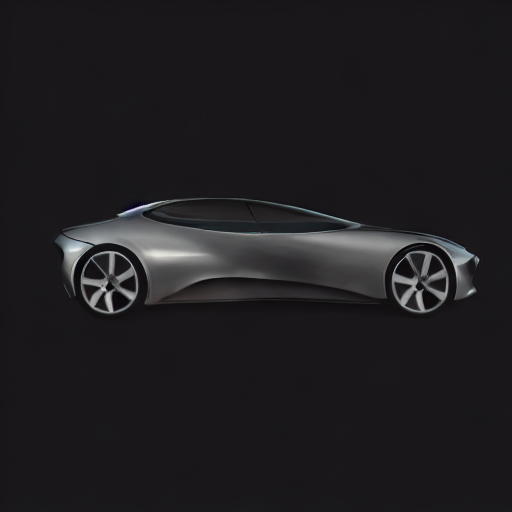} &  \\
    \textbf{Design 2} & \includegraphics[width=0.35\textwidth,trim={0 180px 0 180px},clip]{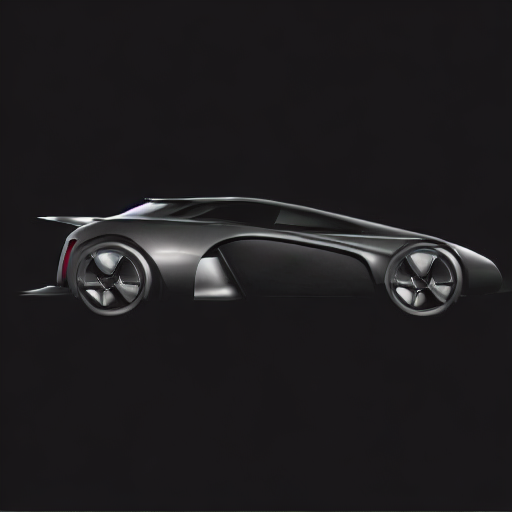} & \\
    \textbf{Design 3} & \includegraphics[width=0.35\textwidth,trim={0 180px 0 180px},clip]{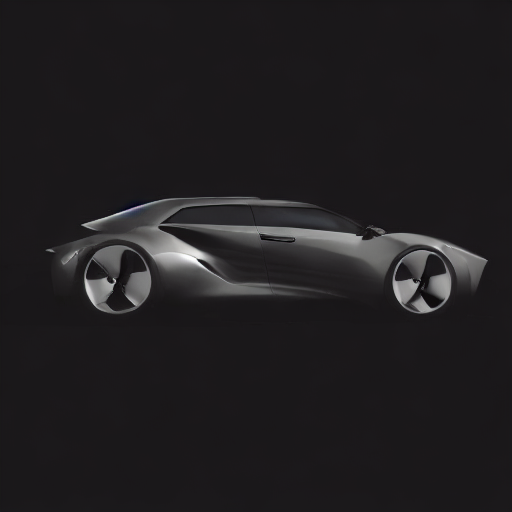} & \\
  \end{tabular}
  \caption{Drag-guided redesigns of a reference image.}\label{fig:drag_redesign}
\end{figure}

Recall that the DDIM sampler is initialized via $x_T = x_{0} + \sigma_T \epsilon$,
where $x_{0}$ denotes a provided reference image and $\epsilon \sim \mathcal{N}(0, I)$.
Deviation of the DDIM output from $x_{0}$ is controlled
by the size of the noise-level $\sigma_T$.  For $\sigma_T$ sufficiently small,
the deviation preserves the basic structure of $x_0$.  This leads to drag-guided \emph{redesigns}  of the reference.
\Cref{fig:drag_redesign} illustrates redesigns
produced by this procedure.  Note that for each
redesign, the initial predicted drag is the same and
equal to the predicted drag of the reference  image.

\subsection{Drag-guided diffusion with a gradient-estimation sampler}\label{sec:gesched}
\begin{figure}
  \centering
  \begin{tabular}{ccc}
    \textbf{Baseline} & \textbf{Drag optimized (DDIM)} & \textbf{Drag optimized (GE)} \\[0.3cm]
    \includegraphics[scale=0.23]{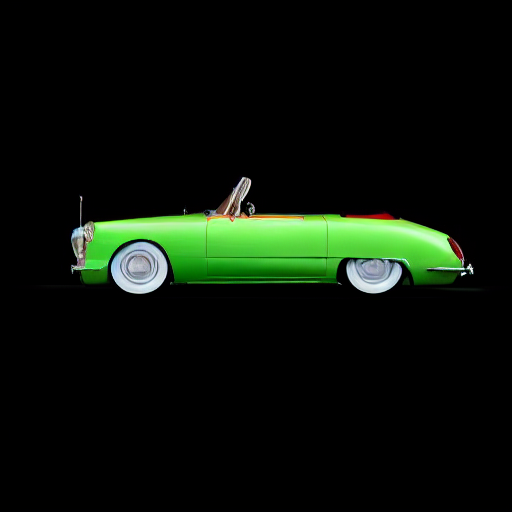} &
    \includegraphics[scale=0.23]{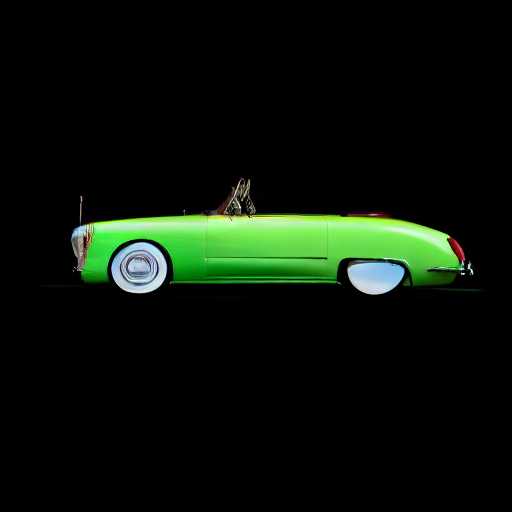}  &
    \includegraphics[scale=0.23]{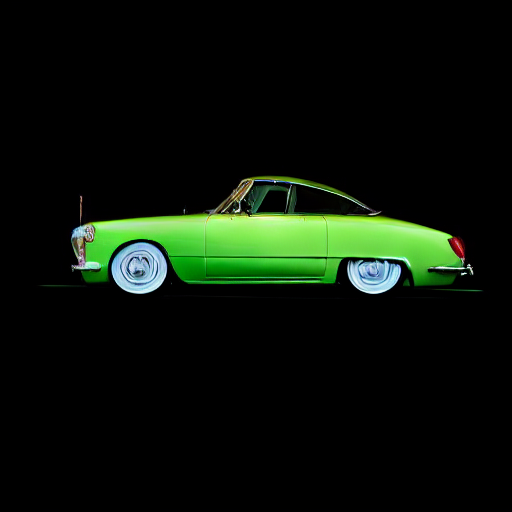}\\
    \includegraphics[scale=0.23]{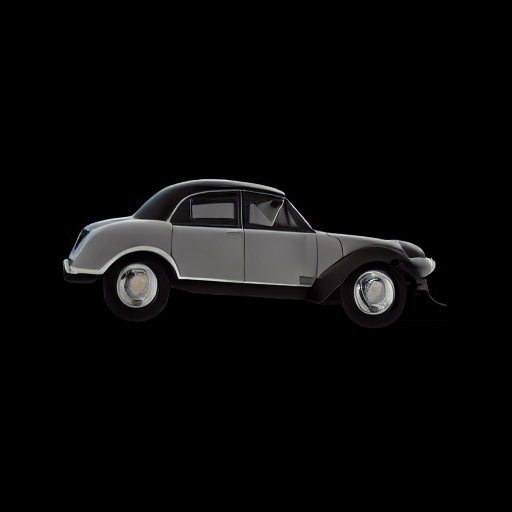} &
    \includegraphics[scale=0.23]{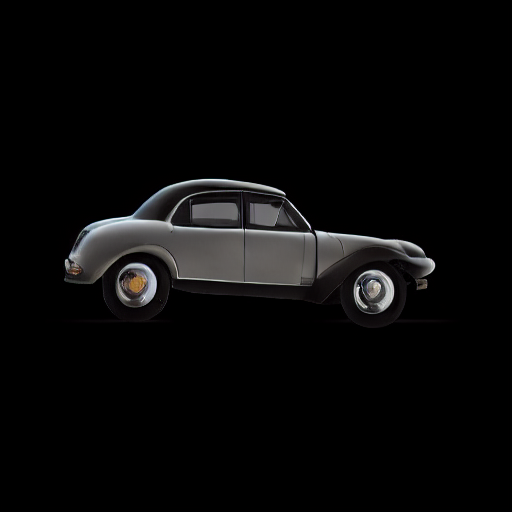}  &
    \includegraphics[scale=0.23]{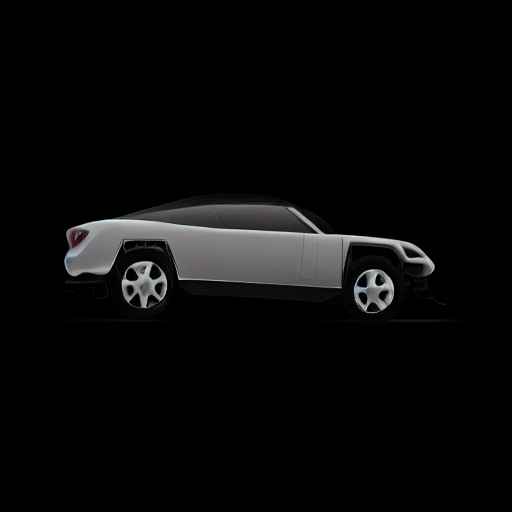}\\
    \includegraphics[scale=0.23]{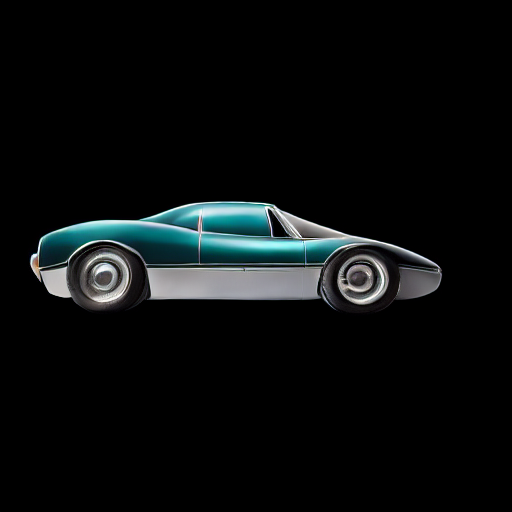} &
    \includegraphics[scale=0.23]{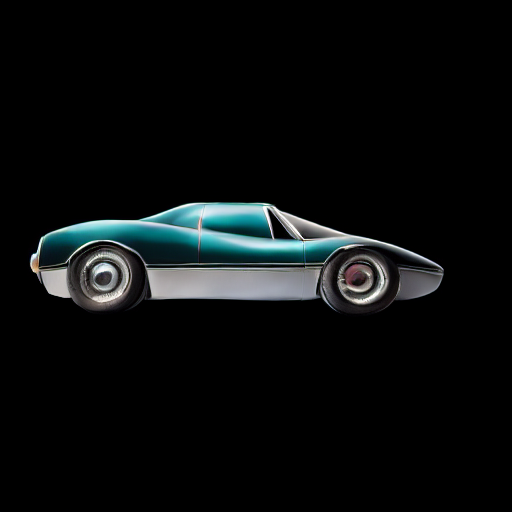}  &
    \includegraphics[scale=0.23]{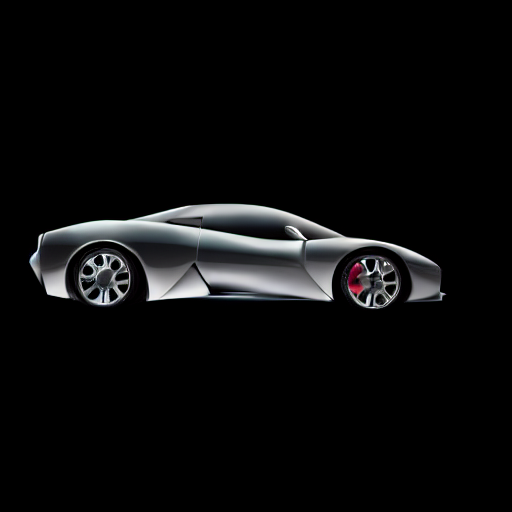}
  \end{tabular}
  \caption{Comparison of DDIM and the gradient-estimation (GE) sampler}~\label{fig:outputs_hos}
\end{figure}

A recent variation on DDIM
is based on reinterpreting sampling
 as approximate
gradient-descent on the Euclidean distance function of the training set~\cite{permenter2023interpreting}.
This leads to a gradient-estimation (GE)
sampler that averages $\epsilon_\theta(x_t, t)$ over multiple time-steps,
exploiting the fact that the gradient of the distance function is constant along
gradient-descent trajectories~\cite[Section 4]{permenter2023interpreting}.
To use this sampler, one replaces $\epsilon_\theta(x_t, t)$ in \eqref{eq:conditioning}
with $\gamma \epsilon_\theta(x_t, t) + (1-\gamma) \epsilon_\theta(x_{t+1}, t+1)]$
for chosen $\gamma \in \mathbb{R}$. We
compare outputs between GE and DDIM samplers in  \Cref{fig:outputs_hos}
using $\gamma = 2$.

\subsection{Distributional robustness of drag estimator}
Our drag estimator $\phi_\text{drag}$ should
generalize to the set of intermediate
denoised images $\hat{x}_0^t := x_t - \sigma_t \epsilon_\theta(x_t, t)$
generated by the diffusion process. There are two
forms of distributional shift that inhibit good generalization. First, the dataset used to
train the denoiser $\epsilon_\theta$ differs from the
dataset used to train $\phi_\text{drag}$.
Second, the distribution of the denoised
outputs $\hat{x}_0^t$ differs from that of the $\epsilon_\theta$-training set,
given that the denoiser is imperfect, i.e., given that $\epsilon \ne \epsilon_\theta(x + \sigma_t \epsilon, t)$.
While it is difficult to evaluate this first type of distributional shift,
 robustness to the second type can be evaluated with controlled
experiments. We carry out two such experiments to evaluate the robustness of $\phi_\text{drag}$
as a function of the chosen feature extractor  (\Cref{sec:drag_estimator}).
We also note that the first type of shift may be mitigated
by retraining $\phi_\text{drag}$ using vehicle images from the Stable
Diffusion training set, but defer this to future research.

\begin{figure}
  \centering
  \begin{tabular}{cc}
    \includegraphics[width=0.35\textwidth]{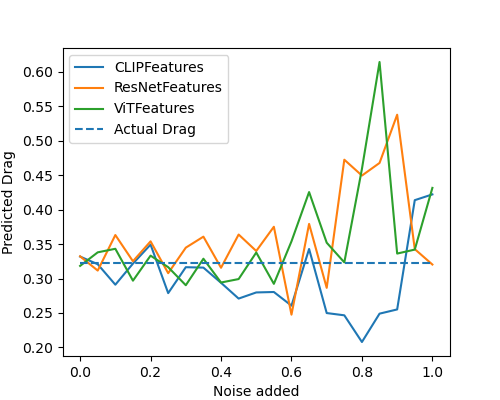} &
    \includegraphics[width=0.45\textwidth,trim={3cm 0 2.5cm 0},clip]{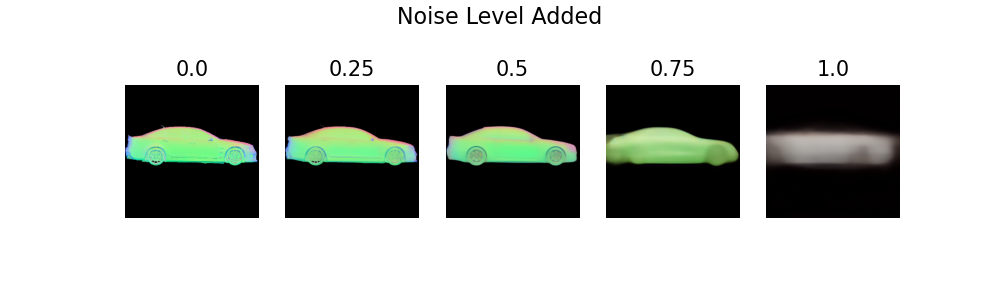}\\
  \end{tabular}
  \caption{Drag estimator performance on out of distribution data: $\hat{x}_0^t$
    with varying amount of noise added to a single image.
  }
  \label{fig:add-noise}
\end{figure}

\begin{figure}
  \centering
  \begin{tabular}{cc}
    \includegraphics[width=0.45\textwidth]{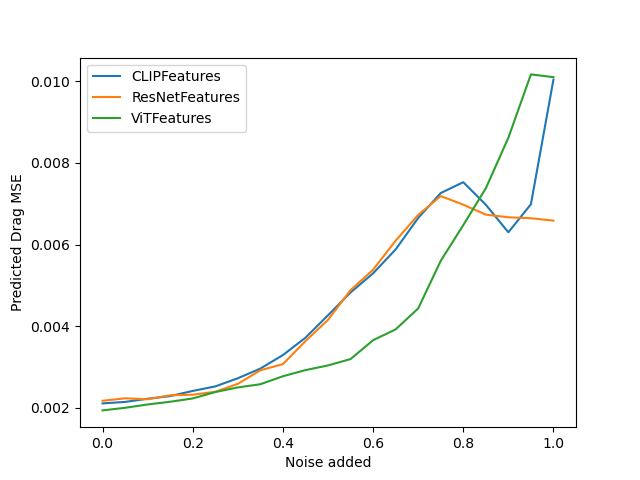} &
    \includegraphics[width=0.45\textwidth]{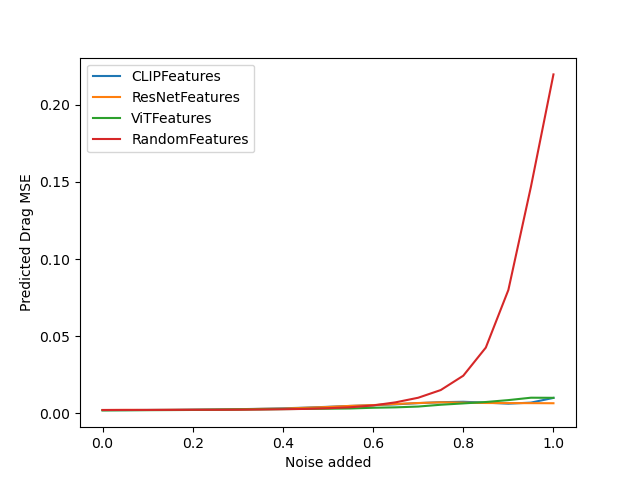}\\
  \end{tabular}
  \caption{Drag estimator performance on out of distribution data as measured by mean-squared
    error on the entire dataset. The left plot is a ``zoomed-in'' version
    of the right plot, showing that random features leads to a much higher
    generalization error than features from pre-trained neural networks.
    }
  \label{fig:add-noise-avg}
\end{figure}

In our first experiment, we take an image used to train $\phi_\text{drag}$ and
add varying levels of noise.  We then predict the drag of the different denoised
images $\hat{x}_0^t$.  A visualization of $\hat{x}_0^t$ and drag predictions for
different noise levels on a single example image is shown in
\Cref{fig:add-noise}. We see that the drag estimator can over or underestimate
the predicted drag, and that the error increases with the magnitude of added noise.
Next, we perform the same experiment over the entire training dataset, and plot
the results in \Cref{fig:add-noise-avg}. The generalization error is measured
with the predicted mean-squared-error of the drag coefficient, which increases
with the amount of noise added.  We see that the features obtained from pretrained
neural networks generalize much better than the random features, even though
they have similar performance on the training data.

\section{Conclusion, Limitations and Future Work}

Generative AI tools that can synthesize engineering
constraints and performance metrics could rapidly
speed-up the design process.
This work is a step towards building such tools and illustrated, as a proof of
concept, that current image generation techniques can be modified
to optimize aerodynamic drag.  Other performance metrics
that can be inferred from images could be incorporated using appropriate
modifications.

A current limitation of our pipeline is its
difficulty controlling---or even measuring---the error
of the surrogate model on the final
drag-optimized output, since the ground-truth
drag coefficient is not uniquely determined
nor computable from the generated 2D rendering.
Future work could alter our approach in a few ways to obtain a unique ground truth.
For instance, one could formulate the problem completely
in 2D by replacing our surrogate model---which
predicts 3D drag from a 2D rendering---with a model that predicts 2D drag
from a planar car.
Alternatively, one can formulate the problem
completely in 3D by replacing Stable Diffusion
with a 3D generative model.
Another limitation of our approach is its 2D image domain formulation,
which may limit its design applications.
Replacing images with CAD or 3D models
is a natural next step. We note combining parametric 3D-models
with image diffusion models is the central task of \emph{score
distillation sampling}~\cite{poole2022dreamfusion}.
Adding physics-based guidance to this technique
is an interesting topic for future research.

\begin{figure}[h!]
  \centering
  \begin{tabular}{ccc}
    \textbf{Baseline} & \textbf{Drag optimized} & \textbf{Estimated Drag} \\[0.3cm]
    \includegraphics[width=0.25\textwidth]{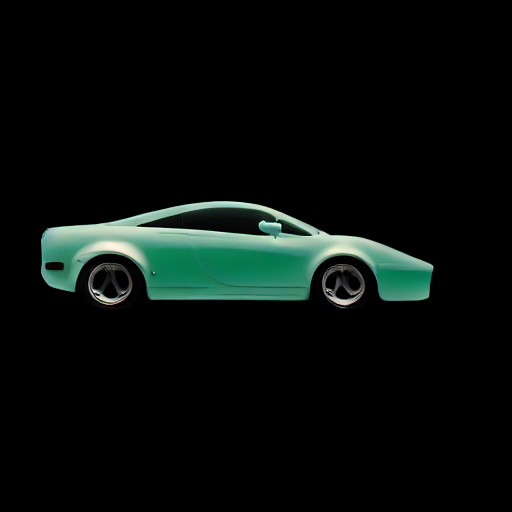}&\includegraphics[width=0.25\textwidth]{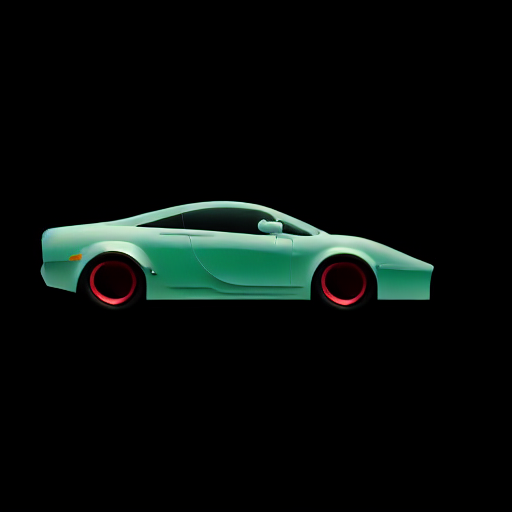}&\includegraphics[width=0.3\textwidth]{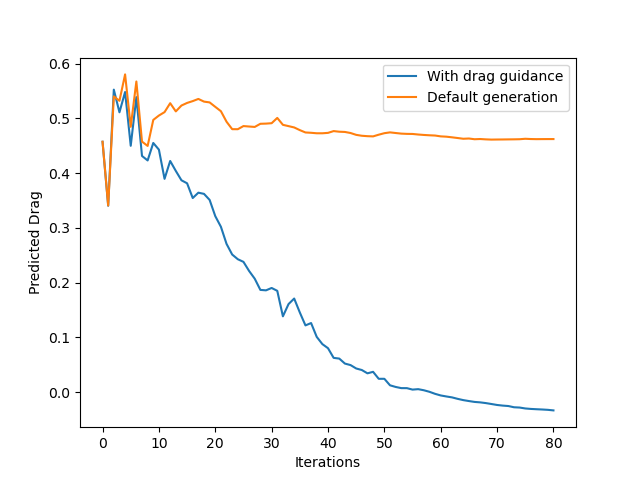}\\
  \end{tabular}
  \caption{Images generated with and without drag guidance
  that demonstrate  incorrect predictions from the surrogate model. While the
  car outline is largely unmodified by guidance,
the surrogate model predicts large improvements and even
negative drag coefficients in the final iterations (blue). }
  \label{fig:outputs_fail}
\end{figure}

Despite unavailability of ground-truth, there are instances
when inaccuracy of the surrogate model is evident.
For example, in Figure~\ref{fig:outputs_fail}, the surrogate model
predicts dramatically different drag coefficients for two cars with
similar outlines. Furthermore, the predicted coefficient for one of the cars is
negative, which is not physically valid. To improve accuracy, future work
could develop active learning
techniques that augment the training set by computing
ground-truth drag coefficients for generated vehicles.  In addition, one could investigate surrogate models
with an implicit physical bias that, for instance,
only return non-negative drag coefficients and always make
similar predictions for similar shapes.

\section*{Acknowledgements}\label{sec:additional_ex}
The authors thank Faez Ahmed for
several useful discussions.

\clearpage

{\small
\bibliographystyle{abbrvnat}
\bibliography{../tex_common/bib}
}

\appendix

\section{Additional examples}\label{sec:additional_ex}

\begin{figure}[H]
  \centering
  \begin{tabular}{ccc}
    \textbf{Baseline} & \textbf{Drag optimized} & \textbf{Predicted Drag}\\[0.3cm]
    \includegraphics[width=0.25\textwidth]{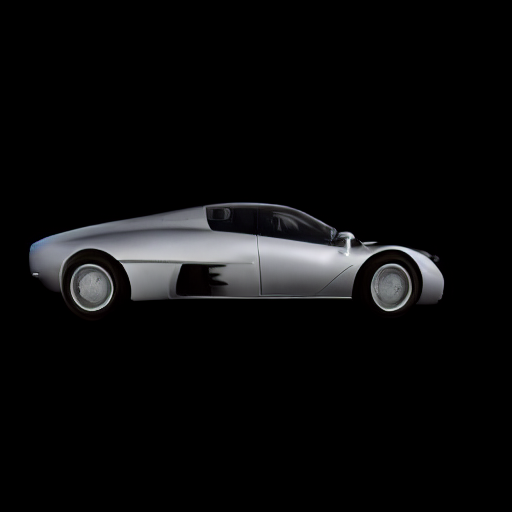}&\includegraphics[width=0.25\textwidth]{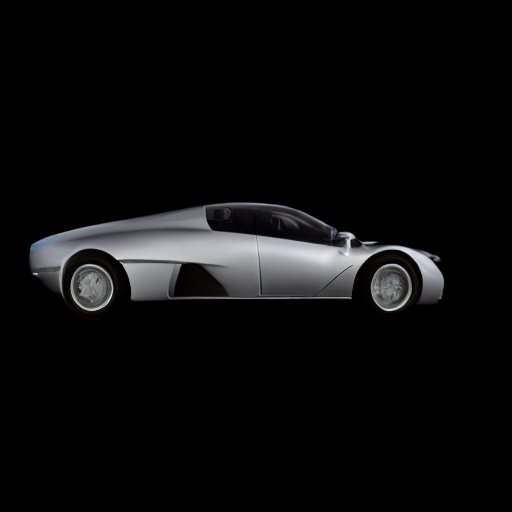}&\includegraphics[width=0.3\textwidth]{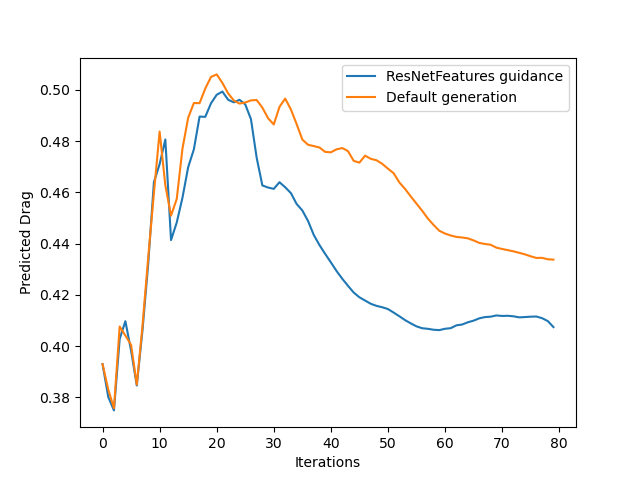}\\ \includegraphics[width=0.25\textwidth]{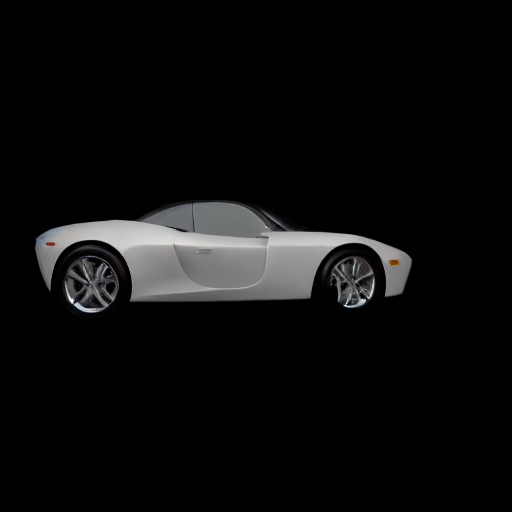}&\includegraphics[width=0.25\textwidth]{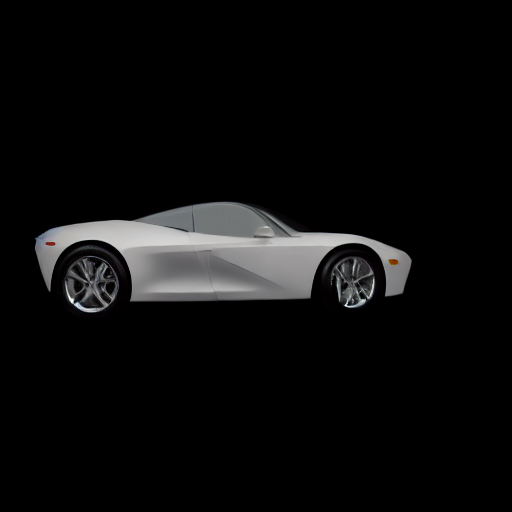}&\includegraphics[width=0.3\textwidth]{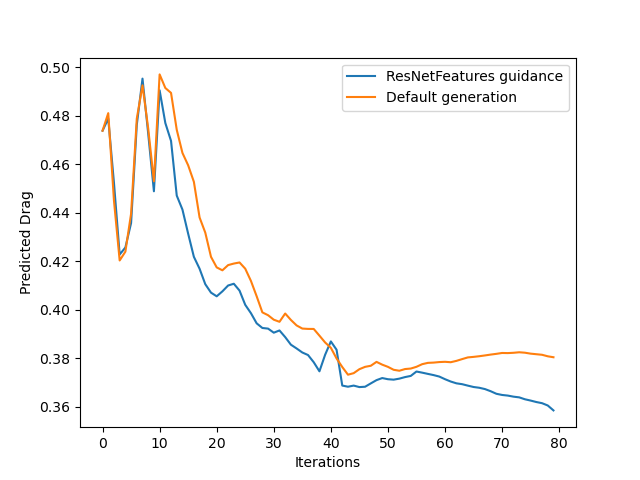}\\ \includegraphics[width=0.25\textwidth]{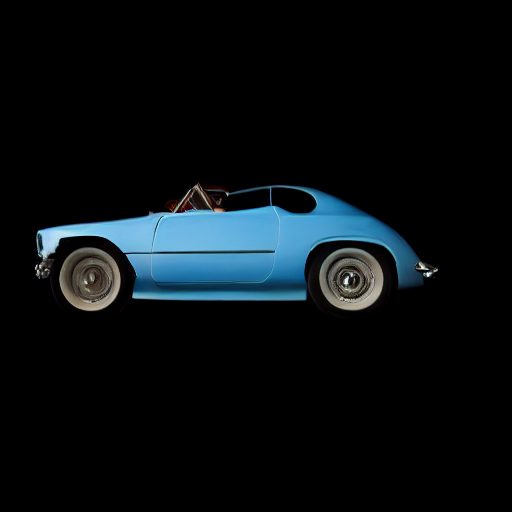}&\includegraphics[width=0.25\textwidth]{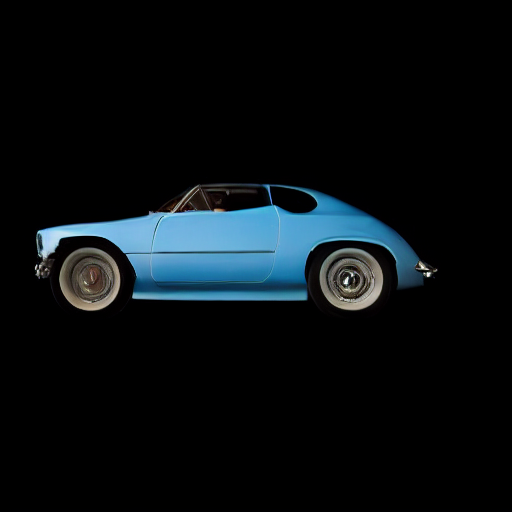}&\includegraphics[width=0.3\textwidth]{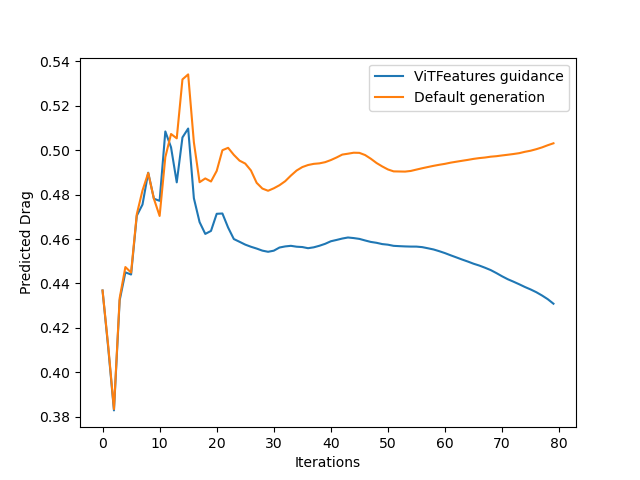}\\ \includegraphics[width=0.25\textwidth]{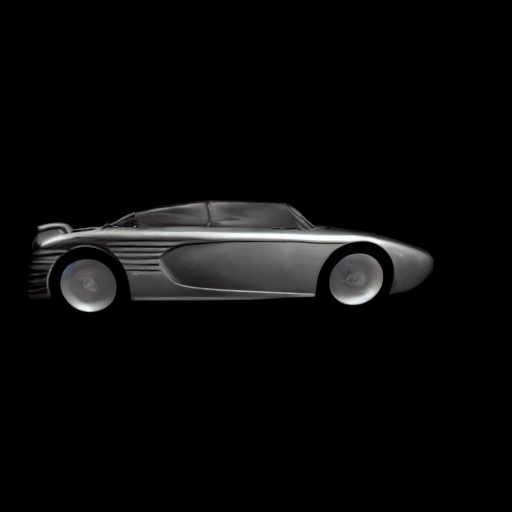}&\includegraphics[width=0.25\textwidth]{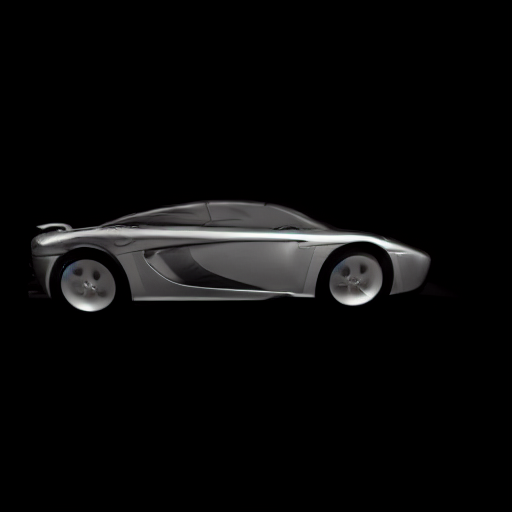}&\includegraphics[width=0.3\textwidth]{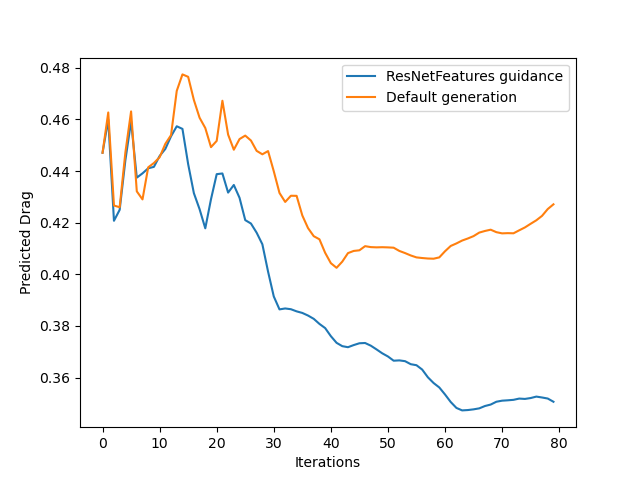}\\ %
  \end{tabular}
  \caption{
Images generated from the prompt ``A car, high resolution studio photo, side view''
with and without drag optimization.
We also plot drag coefficients predicted by the surrogate model.
    }
  \label{fig:outputs-1}
\end{figure}
\begin{figure}[H]
  \centering
  \begin{tabular}{ccc}
    \textbf{Baseline} & \textbf{Drag optimized} & \textbf{Predicted Drag}\\[0.3cm]
      \includegraphics[width=0.25\textwidth]{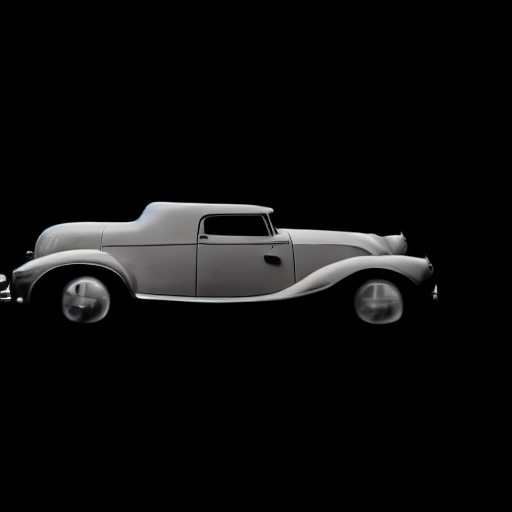}&\includegraphics[width=0.25\textwidth]{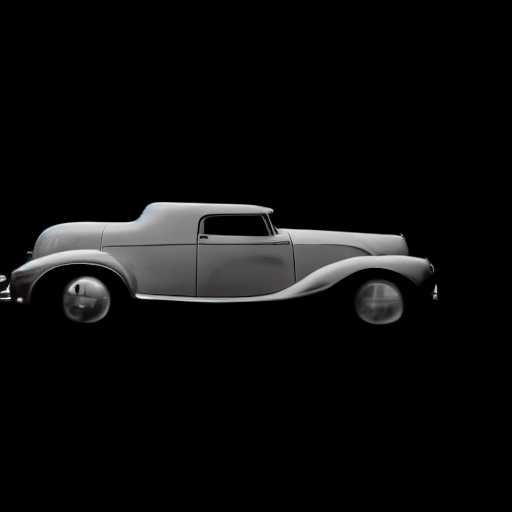}&\includegraphics[width=0.3\textwidth]{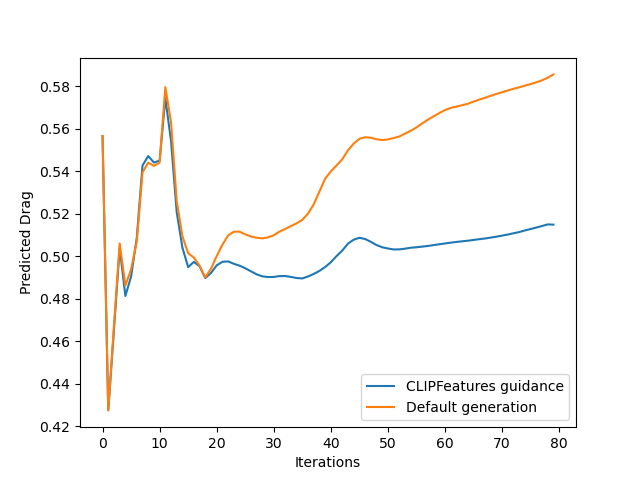}\\ \includegraphics[width=0.25\textwidth]{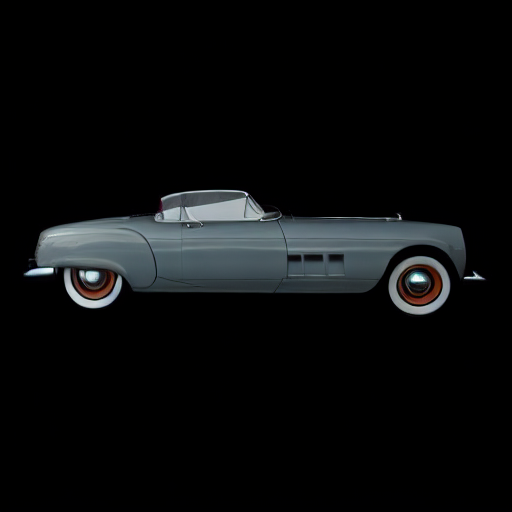}&\includegraphics[width=0.25\textwidth]{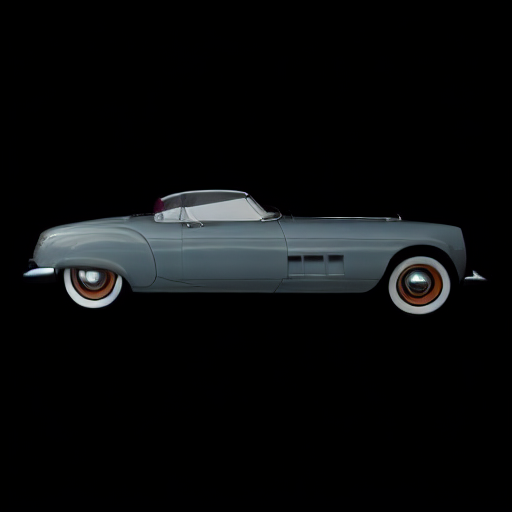}&\includegraphics[width=0.3\textwidth]{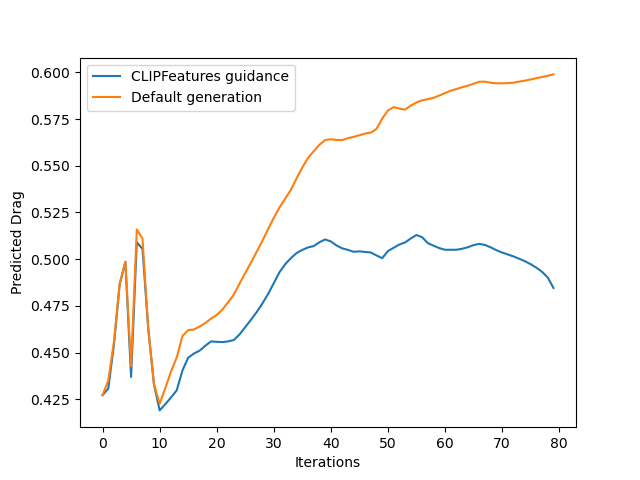}\\ \includegraphics[width=0.25\textwidth]{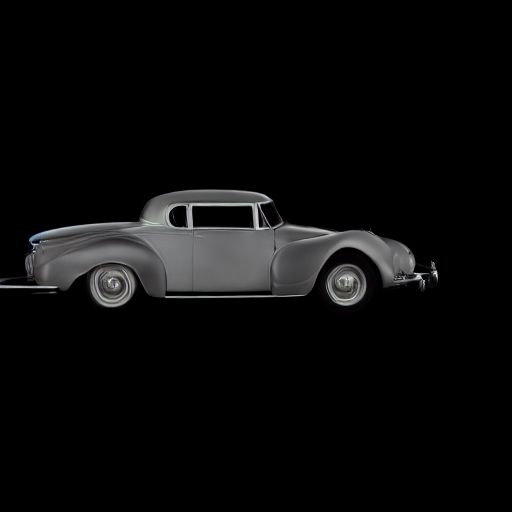}&\includegraphics[width=0.25\textwidth]{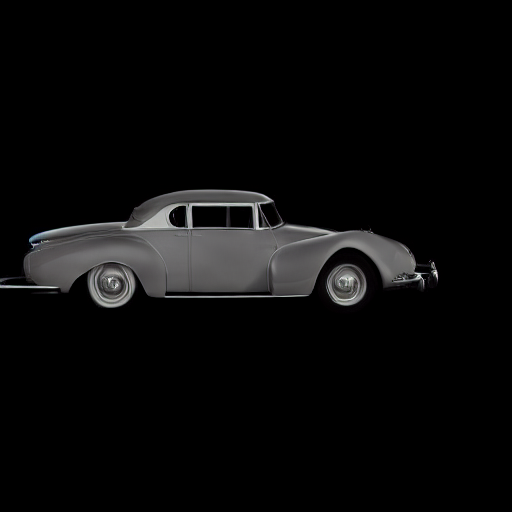}&\includegraphics[width=0.3\textwidth]{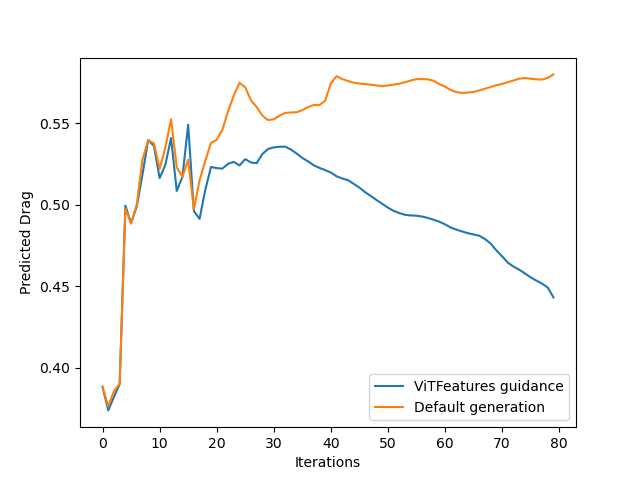}\\ \includegraphics[width=0.25\textwidth]{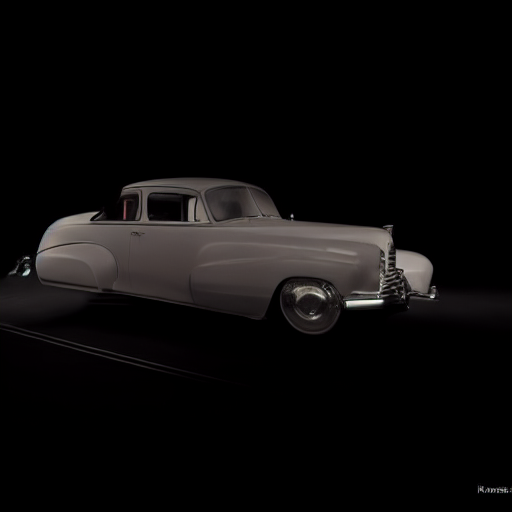}&\includegraphics[width=0.25\textwidth]{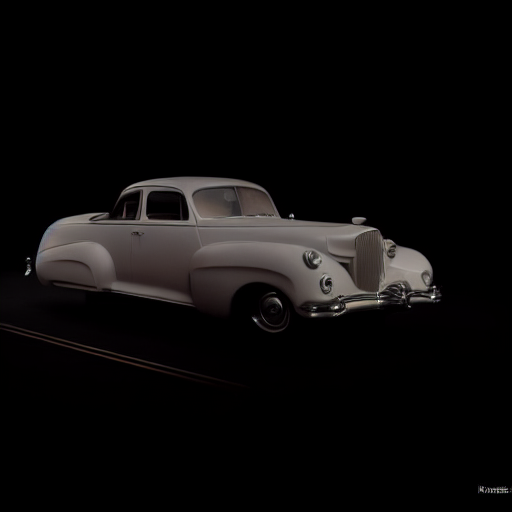}&\includegraphics[width=0.3\textwidth]{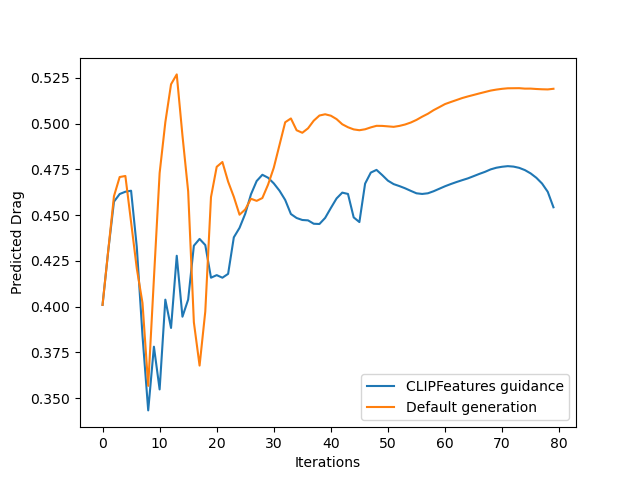}\\
  \end{tabular}
  \caption{
Images generated from the prompt ``An old fashioned car, high resolution studio photo, side view''
with and without drag optimization.
We also plot drag coefficients predicted by the surrogate model.  } \label{fig:outputs-2}
\end{figure}
\begin{figure}[H]
  \centering
  \begin{tabular}{ccc}
    \textbf{Baseline} & \textbf{Drag optimized} & \textbf{Predicted Drag}\\[0.3cm]
    \includegraphics[width=0.25\textwidth]{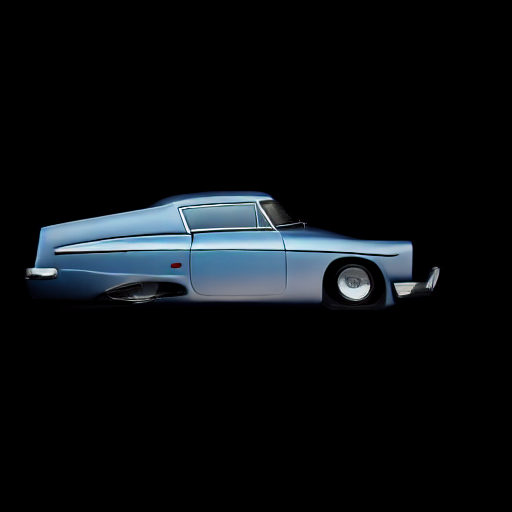}&\includegraphics[width=0.25\textwidth]{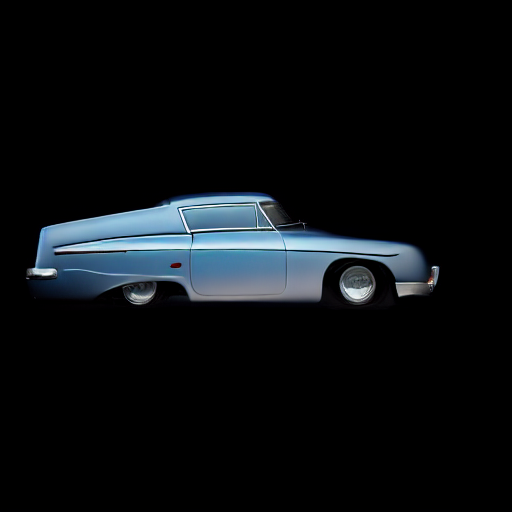}&\includegraphics[width=0.3\textwidth]{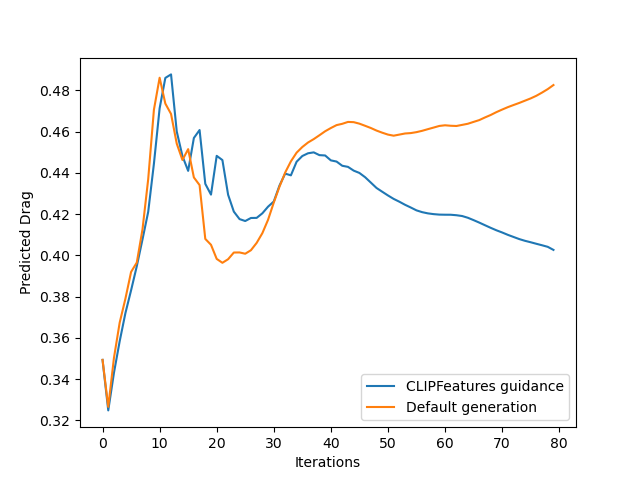}\\ \includegraphics[width=0.25\textwidth]{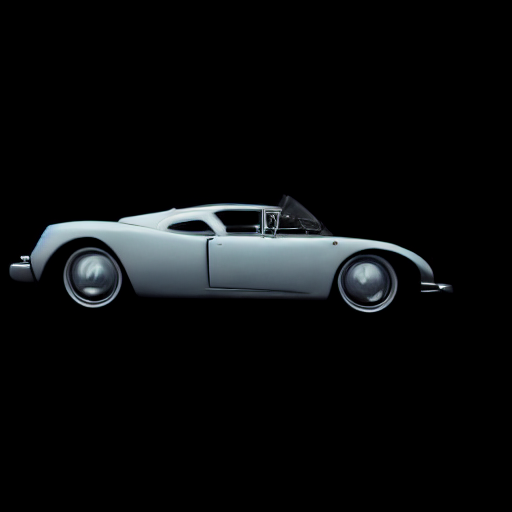}&\includegraphics[width=0.25\textwidth]{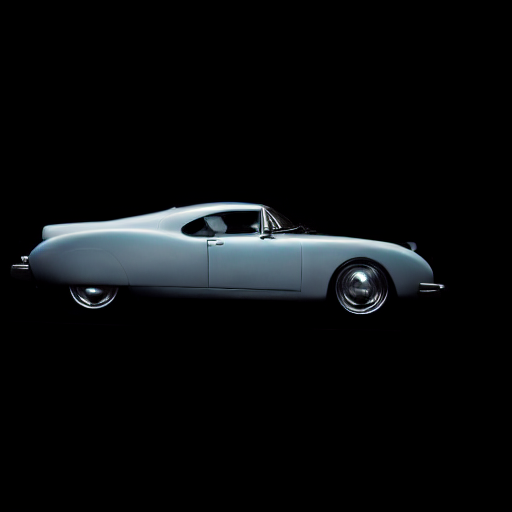}&\includegraphics[width=0.3\textwidth]{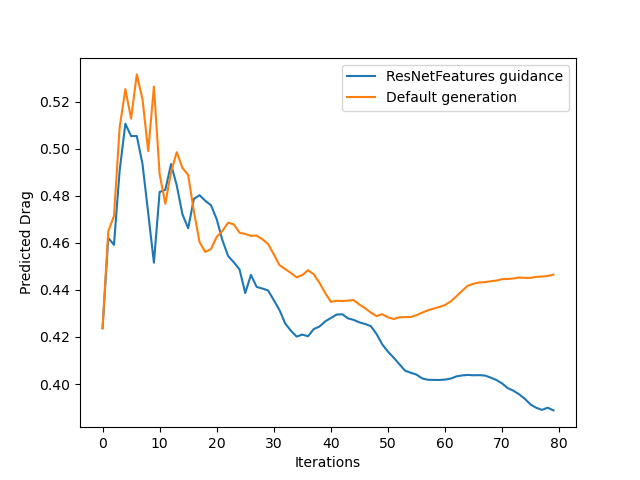}\\ \includegraphics[width=0.25\textwidth]{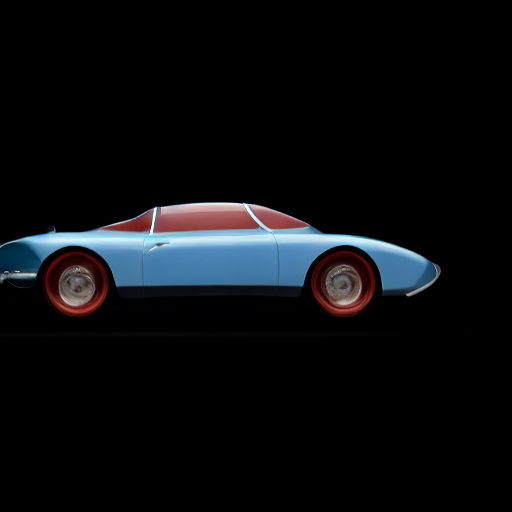}&\includegraphics[width=0.25\textwidth]{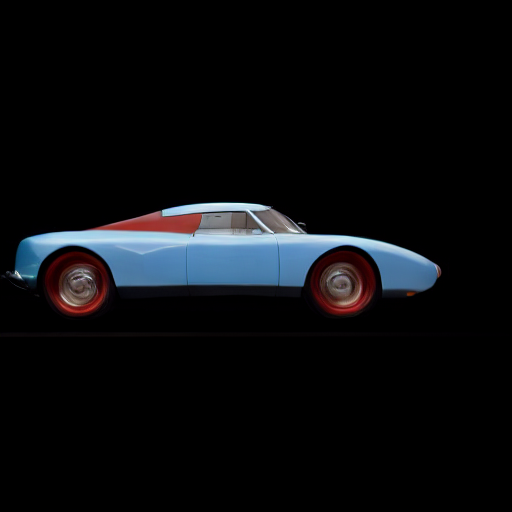}&\includegraphics[width=0.3\textwidth]{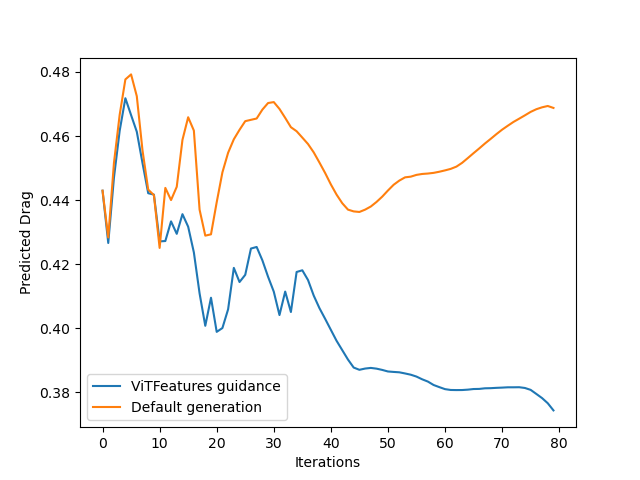}\\ \includegraphics[width=0.25\textwidth]{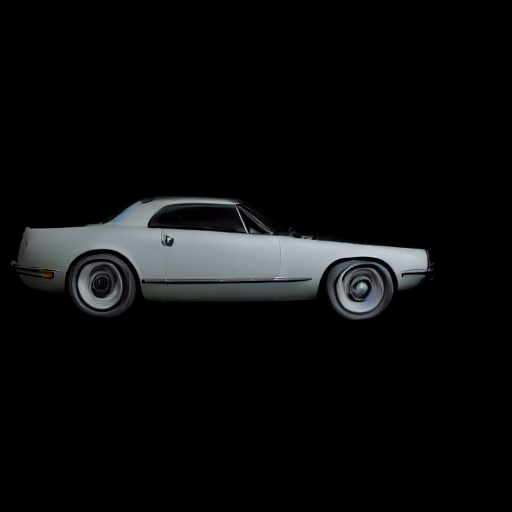}&\includegraphics[width=0.25\textwidth]{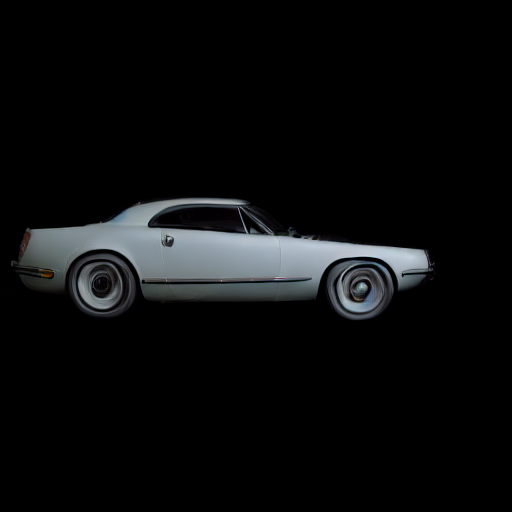}&\includegraphics[width=0.3\textwidth]{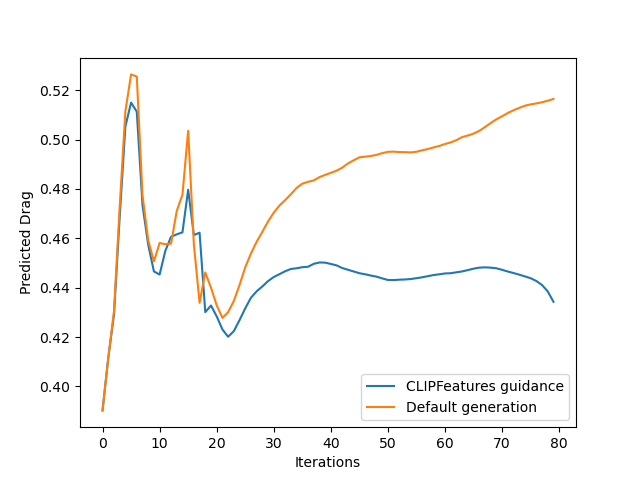}\\ %
  \end{tabular}
  \caption{ Images generated from the prompt ``An iconic car, high resolution studio photo, side view''
with and without drag optimization.
We also plot drag coefficients predicted by the surrogate model.
}~\label{fig:outputs-3}
\end{figure}

\end{document}